\documentclass[10pt,twocolumn,letterpaper]{article}

\usepackage[pagenumbers]{cvpr} 

\usepackage{graphicx}
\usepackage{amsmath}
\usepackage{amssymb}
\usepackage{booktabs}
\usepackage{overpic}
\usepackage{soul}
\usepackage{float}
\usepackage{hyperref}

\usepackage[dvipsnames]{xcolor}

\usepackage{hyperref}
\usepackage{url}
\usepackage{fourier} 
\usepackage{array}
\usepackage{makecell}
\usepackage{graphicx}
\usepackage{array, multirow, bigdelim, makecell, booktabs} 
\usepackage{float}
\usepackage{booktabs}

\usepackage[capitalize]{cleveref}
\crefname{section}{Sec.}{Secs.}
\Crefname{section}{Section}{Sections}
\Crefname{table}{Table}{Tables}
\crefname{table}{Tab.}{Tabs.}

\def\cvprPaperID{10219} 
\def\confName{CVPR}
\def\confYear{2022}

\begin{document}

%
\title{Locally Shifted Attention With Early Global Integration} 

\author{Shelly Sheynin$^{1,2}$\quad Sagie Benaim$^1$ \quad Adam Polyak$^{1,2}$ \quad \quad Lior Wolf$^1$ \\ \\
$^1$Tel Aviv University\quad $^2$Facebook AI Research\\}

\maketitle

\newcommand{\sagie}[1]{\textcolor{blue}{(Sagie: #1)}}
\newcommand{\shelly}[1]{{\textcolor{Magenta}{[Shelly: #1]}}}
\newcommand{\lior}[1]{\textcolor{red}{(Lior: #1)}}
\newcommand{\adam}[1]{\textcolor{green}{(Adam: #1)}}

\begin{abstract}
Recent work has shown the potential of transformers for computer vision applications.
An image is first partitioned into patches, which are then used as input tokens for the attention mechanism. Due to the expensive quadratic cost of the attention mechanism, either a large patch size is used, resulting in coarse-grained global interactions, or alternatively, attention is applied only on a local region of the image, at the expense of long-range interactions. In this work, we propose an approach that allows for both coarse global interactions and fine-grained local interactions already at early layers of a vision transformer. 

At the core of our method is the application of local and global attention layers.  In the local attention layer, we apply attention to each patch and its local shifts, resulting in virtually located local patches, which are not bound to a single, specific location. These virtually located patches are then used in a global attention layer. The separation of the attention layer into local and global counterparts allows for a low computational cost in the number of patches, while still supporting data-dependent localization already at the first layer, as opposed to the static positioning in other visual transformers.   
Our method is shown to be superior to both convolutional and transformer-based methods for image classification on CIFAR10, CIFAR100, and ImageNet. Code is available at: \url{https://github.com/shellysheynin/Locally-SAG-Transformer}
\end{abstract}


\section{Introduction}
\label{sec:intro}

Convolutional neural networks have dominated computer vision research and enabled significant breakthroughs in solving many visual tasks, such as image classification~\cite{krizhevsky2012imagenet,simonyan2014very} and semantic segmentation~\cite{long2015fully}.  
Typically, CNN architectures begin by applying convolutional layers of a small receptive field for low-level features, resulting in local dependencies between neighbouring image regions. As processing continues and features become more semantic, the effective receptive field is gradually increased, capturing longer-ranged dependencies. 

Inspired by the success of Transformers~\cite{vaswani2017attention} for NLP tasks, a new set of attention-based approaches has emerged for vision-based processing. The Vision Transformer (ViT)~\cite{dosovitskiy2020image} is the first model to rely exclusively on the Transformer architecture for obtaining competitive image classification performance. ViT divides the input image into patches of a fixed size and considers each patch as a token to which the transformer model is applied. The attention mechanism between these patches results in global dependencies between pixels, already at the first transformer layer.  
Due to the quadratic cost of the attention mechanism~\cite{vaswani2017attention, dosovitskiy2020image} in the number of patches, fixed-size partitioning is performed. 
 As a result, ViT does not benefit from the built-in locality bias that is present in CNNs: neighbouring pixels within a patch may be highly correlated, but this bias is not encoded into the ViT architecture. That is, ViT encodes inter-patch correlations well, but not intra-patch correlations. 
Further, each image may require a different patch size and location, depending on the size and location of objects in the image.

\begin{table*}
\centering

\begin{tabular}{c|c|c|c|c}
\toprule
Method & Token Embedding &  Hierarchy &  Early Layer Attention &  Late Layer Attention \\
\midrule
ViT~\cite{dosovitskiy2020image} & Linear (non-overlapping) & No & Global & Global \\
DeiT~\cite{touvron2020deit} & Linear (non-overlapping) &  No & Global & Global \\ 
PvT~\cite{wang2021pyramid} & Linear (non-overlapping) & Yes & Global & Global \\
CvT$^*$~\cite{wu2021cvt} & Partial Convolution$^*$ (overlapping) &  Yes & Partially Local + Global$^*$ & Partially Local + Global$^*$ \\
NesT$^{\dagger}$~\cite{zhang2021aggregating}  & Convolution (non-overlapping)  &Yes &  Local & Local + Global$^\dagger$ \\ 
Swin$^{\dagger}$~\cite{ze2021swin}  & Linear (non-overlapping)  &Yes&  Local & Local + Global$^\dagger$ \\
ViT$_C$~\cite{xiao2021early} & Convolution (overlapping) & No & Global & Global \\ 
Ours &  Convolution (overlapping) & Yes & Local + Global &  Global  \\
\bottomrule
\end{tabular}
\caption{A comparison of our method with baselines on key elements. 
We consider the \textit{Token Embedding} in terms of the type of embedding and whether overlapping patches are considered tokens. Next, we consider whether a \textit{Hierarchy} is used, in which a pyramid of attention layers is applied to decreasing input resolutions. We then consider the type of attention used. We divide this into \textit{Early Layer Attention}, in which low-level features are learned, and \textit{Late Layer Attention}, in which higher-level features are learned. 
CvT$^{*}$ applies only partial convolution, which includes depth-wise separable convolution with a stride of $2$ (as opposed to $1$) for the key and value projection. As such, only part of the local neighbourhood of patches is considered for both early and late layers.
In NesT$^{\dagger}$ and Swin$^{\dagger}$, only a local neighborhood of patches is considered, but as the hierarchy progresses, coarser patches are considered, so a local neighborhood covers almost the entire spatial dimension, resulting in a global-like attention. Our method is unique in that it applies fully local and global attention already at the first layer, with full convolutional embedding over the entire local neighborhood of patches. This is possible due to the separation of the attention into local and global components, which reduces overall complexity.
\label{tab:comparison}
}
\end{table*}

Recent approaches~\cite{ze2021swin, zhang2021aggregating, wang2021pyramid, wu2021cvt} have attempted to alleviate the need for this fixed-size partition, thus enjoying some of the benefits of CNNs. PvT~\cite{wang2021pyramid}, for instance, applies attention in a pyramidal fashion, with increasing patch size at each level. However, an initial fixed partition of the image into non-overlapping patches is still performed, so finer sub-patch correlations are still not captured. In another line of work NesT~\cite{zhang2021aggregating} and Swin~\cite{wang2021pyramid}, apply attention in a localized fashion, over local regions in the image. This results in the inability to capture global correlations between distant patches in the image. CvT~\cite{wu2021cvt} considers overlapping patches, thus capturing both inter-patch correlations and intra-patch correlations.
As a result, a large number of patches is considered, and CvT does not scale well to large images due to the quadratic cost of the attention mechanism in the number patches.

Our approach combines the locality bias of CNNs, for both coarser and finer details, with the ability to attend globally to all patches in the image. This is done already at the first layer, for low-level features. The method scales well to large images, since it does not incur the prohibitive quadratic cost of considering all overlapping patches. It is based on the observation that the optimal location for each patch varies from image to image, depending on object locations and sizes. Therefore, instead of considering a single patch at a given location, we consider an ensemble of patches. This ensemble consists of the conventional fixed patch location and of the patches obtained by small horizontal and vertical shifts of each patch. By employing this shift property, the ensemble can capture more precisely the finer details of the object patches, which are necessary for the downstream task. 

To avoid the expensive quadratic cost of computing self-attention over all ensembles of patches, we split each attention layer into two consecutive attention operations, which accumulate both local and global information for each patch. In the local attention layer, we apply self-attention to each patch with its local shifts. This step allows the fixed patches to gain information from a rich collection of variations, where each variant represents an alternative to the location of the patch. This way, we construct a virtually located patch as a weighted sum of all possible shifts of the fixed patch. In the global attention layers, we utilize the virtually local patches and apply the standard global self-attention between them. This step allows each patch to gain global information from all other patches, where each patch was optimized in the previous local layer by considering all of its local shifts. Global attention is applied in a number of layers, in a pyramidal fashion, where at each layer a coarser resolution is considered. 

Our method obtains state-of-the-art performance in image classification on a variety of different model sizes for both CIFAR10, CIFAR100~\cite{krizhevsky2009learning} and ImageNet~\cite{deng2009imagenet}. At the same time, our method is efficient and can scale well to large image resolutions.

\section{Related Work}

\vspace{0.2cm}
\noindent\textbf{Transformers for Vision Models}\quad 
The Transformer, first introduced in~\cite{vaswani2017attention}, revolutionized the field of NLP. 
Multiple attempts have been made to incorporate attention-based techniques for image classification. 
The Vision Transformer (ViT)~\cite{dosovitskiy2020image} marked a turning point for Transformers in vision. ViT showed excellent results compared to existing CNN-based networks, while requiring fewer computational resources.
Later attempts incorporated the locality bias of CNNs within a transformer architecture. DeiT~\cite{touvron2020deit} introduced a teacher-student strategy specifically for Transformers, using an additional distillation token, in which the teacher is a CNN. This enabled training vision Transfromers with the standard ImageNet dataset, removing the need of ViT to utilize a larger-scale pre-training dataset. VIT$_C$~\cite{xiao2021early} showed that applying a convolutional stem on input patches, instead of a linear one, and only then applying standard VIT-like attention blocks, boosts performance. These approaches do not use a hirearchy of attention blocks with decreasing input resolution, and assume a fixed partion of the image intro patches. 

Subsequent approaches introduced such a hierarchy into their design.  PvT~\cite{wang2021pyramid} adopt a pyramidal structure, where the input spatial resolution is decreased, similarly to CNNs, and the patch size is gradually increased. However, an initial fixed partition of the image into non-overlapping patches limits the ability of the attention mechanism to capture finer sub-patch details.

\vspace{0.2cm}
\noindent\textbf{Local and Global Attention}\quad 
Approaches such as the Swin Transformer~\cite{ze2021swin} and NesT~\cite{zhang2021aggregating} attempt to alleviate this issue by focusing on localized self-attention in non-overlapping blocks, and aggregating the blocks. This partition considers very small, even pixel-sized patches, so correlation between surrounding pixels can be considered, thus reinforcing the locality bias of CNNs. 
However, this comes at the expense of considering only a part of the image at a given scale, while ignoring global dependencies between all patches in the image. 

CvT is a recent approach by~\cite{wu2021cvt} that applies the attention mechanism over overlapping patches of the image at different scales, thus capturing finer details along with global dependencies between distant patches. Different overlapping patches are encoded using convolution. However, due to the quadratic cost of the attention mechanism, not all local overlapping patches can be considered. Hence, a stride of 2 (as opposed to 1) is used for the key and value projection. Further, the convolution operation is restricted to be a depth-wise separable convolution, restricting the space of possible solutions. 

Unlike previous methods, our method separates the attention mechanism into a local stage and a global stage, and applies both the local and global components already at the first layer. 
At the local stage, finer details can be aggregated over the local region of each patch, producing new patches, which incorporate fine details. At the global stage, a standard attention computation is applied between all newly aggregated patches. 
Following the first layer, subsequent aggregation is done in a pyramidal fashion over a number of scales, resulting in an efficient transformer architecture, which can scale well to large images, and which incorporates the locality bias at both fine and coarse scales. 

Tab.~\ref{tab:comparison} shows the key differences of our method from previous work. 
Crucially, our method applies non-partial local and global attention at the very first layer. As discussed in Sec.~\ref{sec:method}, assuming that $B$ tokens result from the patch embedding (non-overlapping patches), we can consider a large number of patch variants $T$ (at most $\sqrt{B}$), while the overall complexity remains $O(B^2)$. This results in a much larger number of tokens being used in our attention mechanism, while also considering the entire image region globally.

\section{Method}
\label{sec:method}

We first describe the formation of the patch embeddings. Then, we discuss the two types of attention layers: the local attention layer operates locally on each patch and its shifts,  while the global attention is applied globally, using a hierarchy. Finally, we describe the implementation details. 

\subsection{Local Shift Embeddings}
\label{sec:shifting}

\begin{figure*}[t]
\centering
\begin{tabular}{c}
\includegraphics[height=6cm, width=16cm]{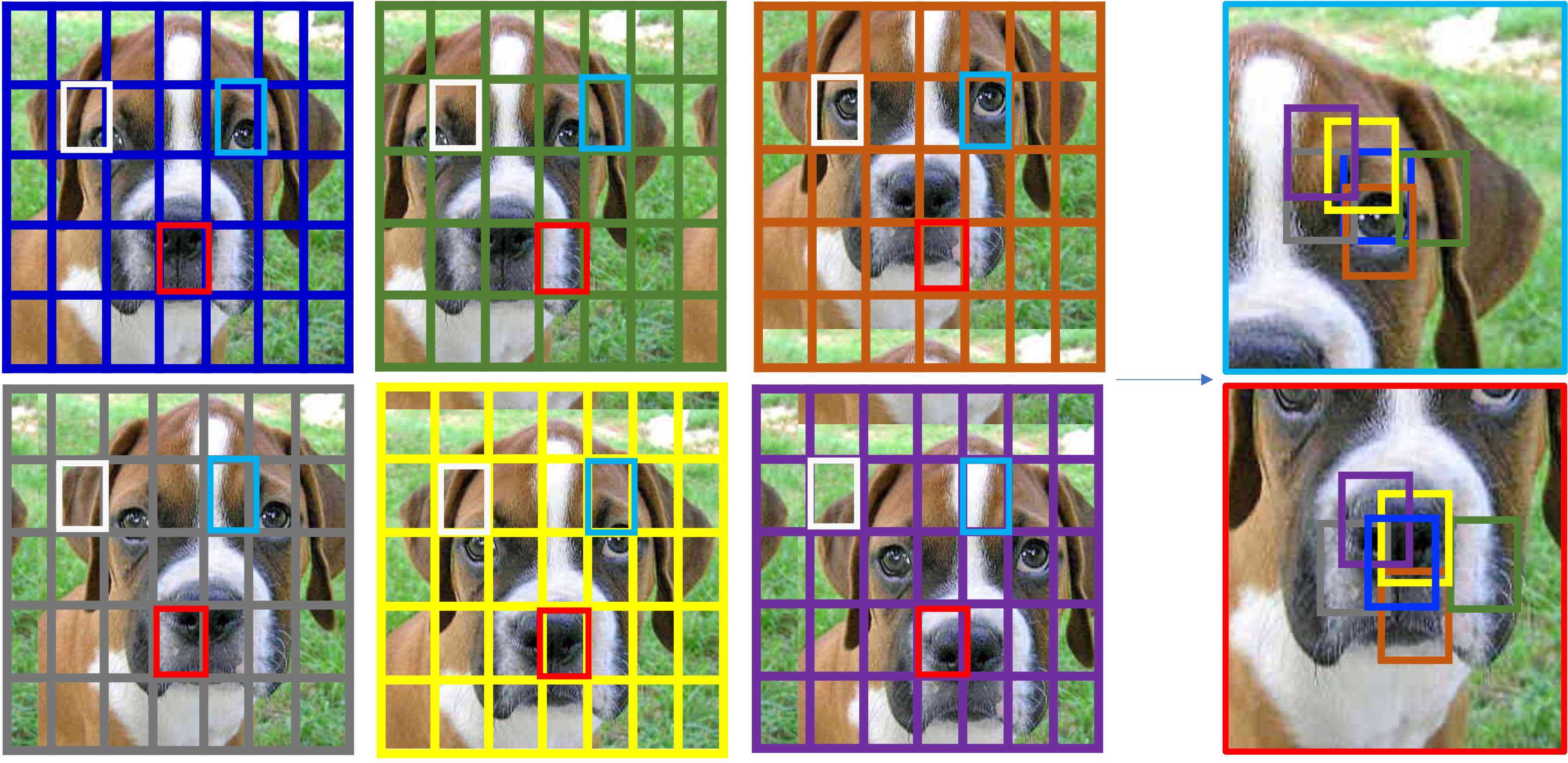} \\
\end{tabular}
\caption{ (a) Illustration of six image shifting variants out of the total $T$, as defined in Sec.~\ref{sec:shifting}. A grid is shown over each image shift variant. In each variant, the grid represents a different division of the image to patches. Patches shown in light blue and red indicate two different patch-shifting variants. The color of the grid is indicative of the variant index. 
(b) The same six variants are shown for the light blue (top) and red  (bottom) patches in (a). The figure illustrates the sub-optimality of the fixed partition used in previous work. For example, the dog's left eye is fully captured only in the "orange partition" and the nose only in the "yellow partition". A fixed partition would only capture one of those concepts. 
\label{fig:shifting_vis} }
\end{figure*}

Given an image $I \in \mathbb{R}^{3 \times H \times W}$, we consider a partition of the image into patches of size $S \times S$, resulting in a map of $\frac{H}{S} \times \frac{W}{S}$ patches, similarly to ViT~\cite{dosovitskiy2020image}. For each patch $i$ of this map, we consider $T$ shifted patches. A patch is created by shifting patch $i$, $P_x$ pixels horizontally and $P_y$ pixels vertically, where $|P_x| \le S$ and $|P_y| \le S$. Each such shift is identified by the pair $(P_x, P_y)$. 

This process results in $T \times \frac{H}{S} \times \frac{W}{S}$ patches of size $S \times S$. The subset of chosen shift variants among all possible shifts is a hyperparameter described in Sec.~\ref{sec:experiments}; In Tab.~\ref{tab:ablation}, we report our results with subsets of varying size.

For each of the $T$ variants of patch $i$, we construct a $D$ dimensional embedding. These embeddings are obtained by shifting the entire image $I$ by $P_x$ and $P_y$, and applying circular padding at the edges. Each resulting image, referred to as an image shifting variant, is then passed through a convolutional layer with $S\times S$ kernel, stride $S$ and $D$ output channels. 
This results in $T$ feature maps of size $D \times \frac{H}{S} \times \frac{W}{S}$. One can view each feature map as $\frac{H}{S} \cdot \frac{W}{S}$ tokens of size $D$. We define $O'^s_i$ to be the token corresponding to patch $i$ and shift index $s$, which corresponds to shift $(P_x,P_y)$. An illustration of our shifting variants is provided in Fig.~\ref{fig:shifting_vis}. 

\vspace{0.2cm}
\noindent\textbf{Positional embedding}\quad
A learned positional embedding is added for each patch and variant convolutional embedding $O'^s_i$. That is, $O^s_i = O'^s_i + L^s_i$, where $L^s_i$ is a learned positional embedding of size $D$. In Sec.~\ref{sec:ablation}, we consider a different strategy, whereby a single $D$ dimensional positional embedding $L_i$is learned for all variants of a given patch, which results in worse performance. 

$O^{s}_i \in \mathbb{R}^D$, for $s=1..T$, is then the set of embeddings for the shift-variants associated with each patch $i$ of size $S\times S$. We assign indexes such that the first variant, $O^1_{i}$, is the embedding of the non-shifted (identity) patch, where $P_x=P_y=0$.

\subsection{Local-global Attention}
\label{sec:localglobal}

We first apply the local attention layer. This attention layer considers each patch $i$ independently. For each $i$, we aggregate the embeddings of the $T$ local shifted variants, using an attention mechanism. This results in an updated embedding of size $D$ for each patch. Next, global attention layers are applied using a pyramidal structure.

\begin{figure*}[t]
\centering
\begin{tabular}{c}
\includegraphics[width=0.995\linewidth]{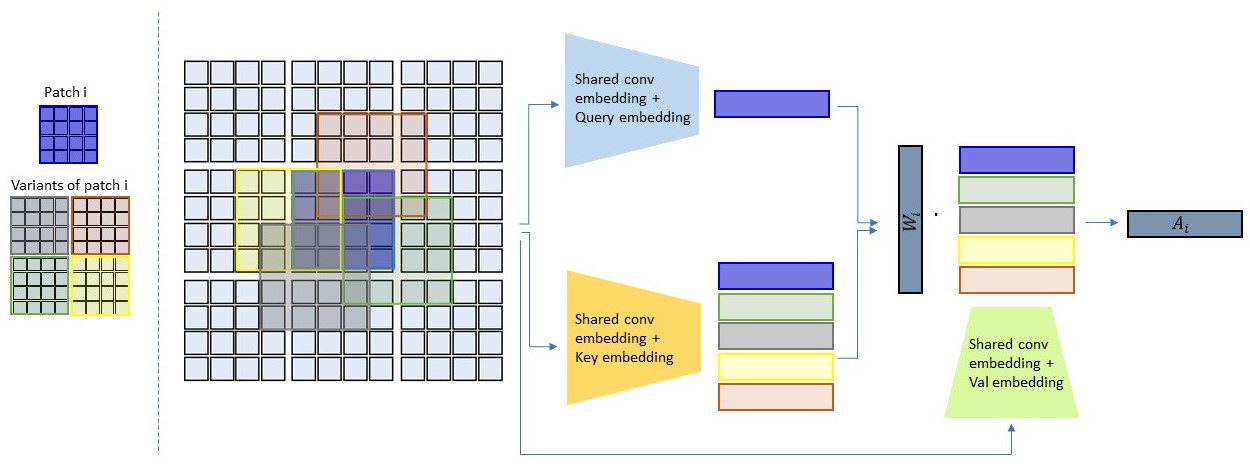} \\
\end{tabular}

\caption{Illustration of our local attention layer. Patch $i$ is shown in blue; variants of $i$ are shown in different colors. All patches are first encoded using a convolutional layer, which, together with the positional encoding, results in a $D$  dimensional embedding $O^{s}_i$ for each patch $i$ and variant $s$. Given $O^{s}_i$, a query embedding is generated as in Eq.~\ref{eq:1}. For each patch variant, a key embedding is generated as in   Eq.~\ref{eq:2}. Both the query and key embeddings are used to create $W_i$ using Eq.~\ref{eq:3}, a softmax probability vector indicating the importance of each patch variant. Finally, $W_i$ is multiplied by the value embeddings (generated using Eq.~\ref{eq:2} for all patch variants) as in Eq.~\ref{eq:4}, to generate the final output $A_i$.  
\label{fig:illustration}} 
\end{figure*}

\vspace{0.2cm}
\noindent\textbf{Local Attention}\quad
An illustration of our local attention layer is shown in Fig.~\ref{fig:illustration}. Denote by $B=\frac{H}{S}\cdot\frac{W}{S}=\frac{HW}{S^2}$ the number of patches for each variant.

The first step is to calculate a query value for each non-shifted patch embedding $O_i^{1}$: 
\begin{align}
 q = XU_{q}  \quad\quad\quad\quad\quad  & X \in \mathbb{R}^{B \times D}, \;\; U_{q} \in \mathbb{R}^{D \times D} \label{eq:1} 
 \end{align}
where $X$ is a matrix constructed from all non-shifted patch embeddings $O_1^{i}$ for $i=1..B$ and $U_{q}$ is a learned query matrix. 
Next, we calculate keys and values for each patch embedding and variant:
\begin{align}
[kv] = ZU_{kv} \quad\quad\quad & Z \in \mathbb{R}^{(TB) \times D}, \;\; U_{kv} \in \mathbb{R}^{D \times 2D} \label{eq:2}
\end{align}
That is, $Z$ is a matrix constructed from all patch embeddings $O_{s}^{i}$  over all variants. $U_{kv}$ is a learned key and value matrix.  We note that, for each patch, while keys and values are computed from all $T$ variants, queries are obtained only from the non-shifted variant.

We now wish to construct an attention matrix $W$. 
We consider each patch separately. That is, given $q \in \mathbb{R}^{B \times D}$ we extract the patch-specific query vector $q_i \in \mathbb{R}^{D}$.
Similarly, we can view $k$ and $v$ as tensors in $\mathbb{R}^{B \times T \times D}$. 
For each patch, we consider the patch-specific key and value matrices $k_i, v_i \in \mathbb{R}^{T \times D}$. We now apply multi-head attention \textit{separately} for each patch:
\begin{align}
W_i = \textit{softmax}(q_ik_i^T/\sqrt{D_h}) \quad\quad &q_i \in \mathbb{R}^{D}, \;\; k_i \in \mathbb{R}^{T \times D} \;\;  \label{eq:3}
\end{align} 
where $D_h$ is the dimension of each attention head. For each patch, this results in a pseudo-probability vector $W_i \in \mathbb{R}^{T}$ indicating the weight of each patch variant. 
The pooled value for patch $i$ is given by:
\begin{align}
A_i = W_iv_i \quad\quad\quad\quad\quad\quad  &W_i \in \mathbb{R}^{T}, \;\; v_i \in \mathbb{R}^{T \times D} \label{eq:4}
\end{align}
$A \in \mathbb{R}^{B \times D} $ is then constructed by pooling all aggregated patch embeddings $A_i$. 

Following the local attention layer, we apply a feed-forward fully-connected
network (FFN) with skip-connection~\cite{he2016deep} and LayerNorm (LN) ~\cite{ba2016layer}.
The application of all of these components together is referred to as a local attention block.

\vspace{0.2cm}
\noindent\textbf{Global Attention}\quad
Following the local attention block, we are given $B$ tokens (for each patch) with an embedding of size $D$. As in the standard setting of transformers \cite{touvron2021training}, a multi-head self-attention~\cite{vaswani2017attention} is applied to the tokens. 
The global attention block consists of a multi-head attention layer followed by a feed-forward fully-connected
network (FFN) with skip-connection~\cite{he2016deep} and LayerNorm (LN) ~\cite{ba2016layer}.

\vspace{0.2cm}
\noindent\textbf{Pyramidal structure}\quad
Recall that $B=\frac{H}{S}\cdot\frac{W}{S}$ , so one can view the input as having a height of $B_h = \frac{H}{S}$, a width of $B_w  = \frac{W}{S}$ and $D$ channels. We consider patches at coarser scales, so we apply a sequence (pyramid) of global attention blocks, but with the output of each block downsampled before applying the next attention block. 

The \textit{downsampling} operation consists of the application of a convolutional layer with a $3\times 3$ kernel, stride $1$ and padding of $1$ followed by a max-pooling operation with a $3\times 3$ kernel, stride $2$ and padding of $1$. Assuming the output channel of the convolutional layer is $D^{down}$, downsampling results in $\frac{B}{4}$ tokens of dimension $D^{down}$, on which the next global attention block is applied. 

We continue this way for $K$ times (chosen as a hyperparameter), obtaining an output with lower and lower resolution. Lastly, we apply global average pooling over the spatial dimension, resulting in a final vector of dimension $D^{final}$. This is followed by a linear layer that outputs $C$ (number of classes) logits, on which standard softmax classification is applied.

\subsection{Implementation Details}
\label{sec:implementation}
In Tab.~\ref{tab:cifar_arch} we provide the exact architectures for two different image resolutions - $224 \times 224$ (used for ImageNet~\cite{deng2009imagenet}) and $32 \times 32$ (used for CIFAR10 and CIFAR100~\cite{krizhevsky2009learning} datasets). For $32 \times 32$ resolution (CIFAR10/CIFAR100) images we consider three model variants provided - tiny, small and base, with increasing number of parameters. For $224 \times 224$ (ImageNet) resolution, tiny and small variants are considered \footnote{Note that while some previous works report results also for larger models, we were unable to allocate the resources needed for such experiments. Specifically, running our larger model on ImageNet would require more than 20 GPU months, using 32GB GPUs. This is not more demanding than previous work. e.g, \cite{wang2021pyramid, ze2021swin, zhang2021aggregating}. However, these resources are not at our disposal at this time.}. 

We use $300$ epochs for all experiments and use the same set of data augmentation and regularization strategies used by \cite{touvron2021training} but exclude repeated augmentations~\cite{hoffer2020augment} and exponential moving average \cite{polyak1992acceleration}. The initial
learning rate is set to $5 \cdot 10^{-4}$. We apply a linear warm-up of $20$ epochs for ImageNet and $5$ epochs for CIFAR10/CIFAR100. We scale the learning rate ($lr$) according to the
batch size ($bs$) as: $\frac{lr}{512} \times bs$. We use the  AdamW~\cite{kingma2014adam} optimizer with a cosine learning rate scheduler. The weight decay is set to $0.05$ and the maximal gradient norm is clipped to $5.0$.

\begin{table*}[t]
\renewcommand{\arraystretch}{1.5}
\centering
\scalebox{0.73}{
\begin{tabular}{|l|l|l|l|l|l|l|lll|}
\toprule
\multicolumn{5}{c|} {Input Resolution: 224 $\times$ 224} & \multicolumn{5}{c} {Input Resolution: 32 $\times$ 32} \\
\cmidrule(r){1-5}
\cmidrule(r){6-10}
 & Output size & Layer & \multicolumn{1}{l|}{Tiny} & \multicolumn{1}{l|}{Small}  & Output size & Layer   & \multicolumn{1}{l|}{Tiny} & \multicolumn{1}{l|}{Small} & Base  \\
\midrule
1  & \begin{tabular}{@{}l@{}} T $\times$ $E_1$ $\times \frac{H}{S} \times \frac{W}{S} $ \\  T $\times (\frac{H}{S} \cdot \frac{W}{S}) \times$ $E_1$  \\ $(\frac{H}{S} \cdot \frac{W}{S}) \times$ $E_1$\\ \end{tabular} & \begin{tabular}{@{}l@{}}  Conv. emb.  \\   Proj. \\ Local Att. \\ \end{tabular}  &   \begin{tabular}{@{}ll@{}} $ K = S = 7 $ & \hspace{-1em}\rdelim\}{2}{*}[${}\times T $] \\ $P = 0 $     \end{tabular}  & \begin{tabular}{@{}ll@{}} $ K = S = 4$ & \hspace{-1em}\rdelim\}{2}{*}[${}\times T $] \\ $P = 0 $     \end{tabular}  &   \begin{tabular}{@{}l@{}} T $\times$ $E_1$ $\times H \times W $ \\  T $\times (H \cdot W) \times$ $E_1$  \\ \end{tabular} & \begin{tabular}{@{}l@{}}  Conv. emb.  \\   Proj. \\ Local Att. \\ \end{tabular} & \begin{tabular}{@{}ll@{}} $K = 3$ &  \hspace{-1em}\rdelim\}{2}{*}[${}\times T $ ]  \\ $S = P = 1$    \end{tabular}   & & \\
\midrule
2     &   \begin{tabular}{@{}l@{}} $(\frac{H}{S} \cdot \frac{W}{S})$ $\times$ $E_1$ \\  $E_2$ $\times$ $ \frac{H}{2S} \times \frac{W}{2S}$ \\   $ (\frac{H}{2S} \cdot \frac{W}{2S})\times$ $E_2$  \\ \end{tabular}     &  \begin{tabular}{@{}l@{}}  Global Att. \\  Down. \\  Proj.  \\ \end{tabular}      &   \multicolumn{1}{l|}{\begin{tabular}{@{}ll@{}} $E_1 = 192 $ & \hspace{-1em}\rdelim\}{2}{*}[${}\times2$ ]  \\ $H_1 = 4$\end{tabular}}   &  \multicolumn{1}{l|}{\begin{tabular}{@{}ll@{}} $E_1 = 64 $ & \hspace{-1em}\rdelim\}{2}{*}[${}\times2$ ]  \\ $H_1 = 2$\end{tabular}} &     \begin{tabular}{@{}l@{}} $(H \cdot W)$ $\times$ $E_1$ \\  $E_2$ $\times$ $ \frac{H}{2} \times \frac{W}{2}$ \\   $ (\frac{H}{2} \cdot \frac{W}{2})\times$ $E_2$  \\ \end{tabular} & \begin{tabular}{@{}l@{}}  Global Att. \\  Down. \\  Proj.  \\ \end{tabular}        & \multicolumn{1}{l|}{\begin{tabular}{@{}ll@{}} $E_1 = 192 $ & \hspace{-1em}\rdelim\}{2}{*}[${}\times2$ ]  \\ $H_1 = 3$\end{tabular}} &  \multicolumn{1}{l|}{\begin{tabular}{@{}ll@{}} $E_1 = 384 $ & \hspace{-1em}\rdelim\}{2}{*}[${}\times2$ ]  \\ $H_1 = 6$\end{tabular}}   &    \multicolumn{1}{l|}{\begin{tabular}{@{}ll@{}} $E_1 = 768 $ & \hspace{-1em}\rdelim\}{2}{*}[${}\times3$ ]  \\ $H_1 = 12$\end{tabular}}   \\ \midrule
3      &    \begin{tabular}{@{}l@{}}  $ (\frac{H}{2S} \cdot \frac{W}{2S}) \times$ $E_2$ \\  $E_3$ $\times \frac{H}{4S} \times \frac{W}{4S}$  \\   $ (\frac{H}{4S} \cdot \frac{W}{4S}) \times$ $E_3$  \\ \end{tabular}    &   \begin{tabular}{@{}l@{}}  Global Att. \\  Down. \\  Proj.  \\ \end{tabular}     &  \multicolumn{1}{l|}{\begin{tabular}{@{}ll@{}}   $E_2 = 192 $ & \hspace{-1em}\rdelim\}{2}{*}[${}\times4$ ]   \\ $H_2 = 4 $\end{tabular}}  & \multicolumn{1}{l|}{\begin{tabular}{@{}ll@{}}   $E_2 = 192 $ & \hspace{-1em}\rdelim\}{2}{*}[${}\times2$ ]   \\ $H_2 = 6 $\end{tabular}}    &      \begin{tabular}{@{}l@{}}  $ (\frac{H}{2} \cdot \frac{W}{2}) \times$ $E_2$ \\  $E_3$ $\times \frac{H}{4} \times \frac{W}{4}$  \\   $ (\frac{H}{4} \cdot \frac{W}{4}) \times$ $E_3$  \\ \end{tabular}        & \begin{tabular}{@{}l@{}}  Global Att.\\  Down. \\  Proj.  \\ \end{tabular}   &  \multicolumn{1}{l|}{\begin{tabular}{@{}ll@{}}   $E_2 = 192 $ & \hspace{-1em}\rdelim\}{2}{*}[${}\times4$ ]   \\ $H_2 = 3 $\end{tabular}}    &     \multicolumn{1}{l|}{\begin{tabular}{@{}ll@{}} $E_2 = 384 $ & \hspace{-1em}\rdelim\}{2}{*}[${}\times4$ ]  \\ $H_2 = 6$\end{tabular}}  &    \multicolumn{1}{l|}{\begin{tabular}{@{}ll@{}} $E_2 = 768 $ & \hspace{-1em}\rdelim\}{2}{*}[${}\times3$ ]  \\ $H_2 = 12$\end{tabular}}      \\ \midrule
4   &     \begin{tabular}{@{}l@{}}   $ (\frac{H}{4S} \cdot \frac{W}{4S}) \times$ $E_3$ \\ $E_4$ $\times \frac{H}{8S} \times \frac{W}{8S}$  \\ $ (\frac{H}{8S} \cdot \frac{W}{8S}) \times$ $E_4$     \\ \end{tabular}   &    \begin{tabular}{@{}l@{}}  Global Att.\\  Down. \\  Prof. \\ \end{tabular}      &  \multicolumn{1}{l|}{\begin{tabular}{@{}ll@{}}   $E_3 = 192 $ & \hspace{-1em}\rdelim\}{2}{*}[${}\times4$ ]   \\ $H_3 = 4 $\end{tabular}}  &  -     &           \begin{tabular}{@{}l@{}}  $ (\frac{H}{4} \cdot \frac{W}{4}) \times$ $E_3$ \\  $E_4$ $\times \frac{H}{8} \times \frac{W}{8}$  \\   $ (\frac{H}{8} \cdot \frac{W}{8}) \times$ $E_4$  \\ \end{tabular}    & \begin{tabular}{@{}l@{}}  Global Att. \\  Down. \\  Proj.  \\ \end{tabular}    &  \multicolumn{1}{l|}{\begin{tabular}{@{}ll@{}}   $E_3 = 192 $ & \hspace{-1em}\rdelim\}{2}{*}[${}\times4$ ]   \\ $H_3 = 3 $\end{tabular}}    &     \multicolumn{1}{l|}{\begin{tabular}{@{}ll@{}} $E_3 = 384 $ & \hspace{-1em}\rdelim\}{2}{*}[${}\times4$ ]  \\ $H_3 = 6$\end{tabular}}  &  \multicolumn{1}{l|}{\begin{tabular}{@{}ll@{}} $E_3 = 768 $ & \hspace{-1em}\rdelim\}{2}{*}[${}\times3$ ]  \\ $H_3 = 12$\end{tabular}}   \\ \midrule

5  &   \begin{tabular}{@{}l@{}}   $ (\frac{H}{F} \cdot \frac{W}{F}) \times$ $E_4$ \\  $E_4$ \\  \# classes  \\ \end{tabular}  &  \begin{tabular}{@{}l@{}}  Global Att.\\  Avg Pool \\  Linear. \\ \end{tabular}  &   \multicolumn{1}{l|}{\begin{tabular}{@{}ll@{}} $F=8S$ \\   $E_4 = 192 $ & \hspace{-1em}\rdelim\}{2}{*}[${}\times4$ ]   \\ $H_4 = 4 $\end{tabular}} & \multicolumn{1}{l|}{\begin{tabular}{@{}ll@{}} $F=4S$ \\$E_4 = 384 $ & \hspace{-1em}\rdelim\}{2}{*}[${}\times10$ ]   \\ $H_4 = 12 $\end{tabular}}  &      \begin{tabular}{@{}l@{}}   $ (\frac{H}{8} \cdot \frac{W}{8}) \times$ $E_4$ \\  $E_4$ \\  \# classes  \\ \end{tabular}   & \begin{tabular}{@{}l@{}}  Global Att. \\  Avg Pool \\  Linear. \\ \end{tabular}  & \multicolumn{1}{l|}{\begin{tabular}{@{}ll@{}} $E_4 = 192 $ & \hspace{-1em}\rdelim\}{2}{*}[${}\times4$ ]   \\ $H_4 = 3 $\end{tabular}}     &   \multicolumn{1}{l|}{\begin{tabular}{@{}ll@{}} $E_4 = 384 $ & \hspace{-1em}\rdelim\}{2}{*}[${}\times4$ ]  \\ $H_4 = 6$\end{tabular}}   &  \multicolumn{1}{l|}{\begin{tabular}{@{}ll@{}} $E_4 = 768 $ & \hspace{-1em}\rdelim\}{2}{*}[${}\times3$ ]  \\ $H_4 = 12$\end{tabular}}   \\ \bottomrule
\end{tabular}}

\caption{Architectures used for images with input resolution $224 \times 224$ (ImageNet) and for input resolution $32 \times 32$ (CIFAR10 and CIFAR100) for three model types: tiny, small and base. In step 1, we begin by applying a convolutional embedding  (Conv emb.) to the input image. A kernel size $K$, stride $S$ and padding $P$ are used and the output has $E_1$ channels. A separate convolutional embedding is performed for each of the $T$ image shifting variants (indicated beside the curly bracket) and positional encoding is added for each patch and variant embedding. Subsequently, we project the image (Proj.) into a three-dimensional tensor, which is then used as input for our local attention block (Local Att.). From step 2 onward, we begin by applying a global attention block (Glob Att.). We subsequently un-project the input and apply a downsampling operation (Down), as described in Sec.~\ref{sec:localglobal}, where the resulting dimension is $E_i$. $H_i$ denotes the number of heads used. A projection layer is then applied, converting the input into a two-dimensional input that can be used for subsequent global attention blocks. Each such step (Global Att. - Down. - Proj.) may be applied a number of times, as indicated by the number beside the curly bracket. In the last step, downsampling is replaced by average pooling (Avg. Pool) and projection is replaced by a linear layer (Linear), resulting in a logit vector. The final step (not shown) is a standard application of softmax, followed by a cross-entropy loss.
\label{tab:cifar_arch} }
\end{table*}

\section{Experiments}
\label{sec:experiments}
We present multiple image classification experiments. Results are reported on three datasets: CIFAR10~\cite{krizhevsky2009learning}, CIFAR100~\cite{krizhevsky2009learning}, and ImageNet~\cite{deng2009imagenet}. Evaluation on the CIFAR10 and CIFAR100 datasets demonstrates the effectiveness of our method on low-resolution $32 \times 32$ images, while evaluation on ImageNet demonstrates the effectiveness of our method on a higher resolution of $224 \times 224$. 

We consider state-of-the-art convolution-based baselines as well as transformer-based baselines. Beyond DeiT~\cite{touvron2020deit}, we also consider baselines that use a pyramidal architecture: PVT~\cite{wang2021pyramid}, Swin~\cite{ze2021swin} and Nest~\cite{zhang2021aggregating}. CvT~\cite{wu2021cvt} is also considered for ImageNet (CIFAR10/CIFAR100 values not reported). Convolutional baselines include  Pyramid-164-48~\cite{han2017deep} and WRN28-10~\cite{zagoruyko2016wide} for CIFAR10/CIFAR100 and ResNet50, ResNet101~\cite{he2016deep}, ResNetY-4GF and ResNetY-8GF~\cite{radosavovic2020designing} for ImageNet.  
We also perform an extensive ablation study for evaluating the contribution of each of our components.

\vspace{0.2cm}
\noindent\textbf{CIFAR10/CIFAR100}\quad
As noted previously by \cite{zhang2021aggregating}, existing transformer-based methods usually perform poorly on such datasets. This is because self-attention methods are typically data-intensive, while in these datasets both the resolution and the number of samples are relatively small. 
In contrast, our method reintroduced the locality bias of CNNs by first applying local attention over neighbouring patches. By subsequently applying global attention in a pyramidal fashion, our method benefits from the introduction of correlations between distant patches gradually, at different scales. {For CIFAR10 and CIFAR100 datasets, we use $T=18$ possible local variants: \{(0,0), (1,0), (0,1), (-1,0), (0,-1), (2,0), (0,2), (-2,0), (0,-2), (1,1), (1,2), (1,-1), (-1,1), (-1,2), (-1,-1), (2,1), (2,2), (2,-1)\}}.

The results for CIFAR10 and CIFAR100 experiments are shown in Table~\ref{tab:cifar}. As can be seen, our method is superior to both CNN-based and Transformer-based baselines for each model size (tiny, small and base).
Already at a small model size of $10.2$M parameters our model achieves superior performance for both CIFAR10 and CIFAR100 on most baselines, which have a significantly larger number of parameters.

\begin{table*}[t]
\centering
\renewcommand{\arraystretch}{0.95}
\setlength{\tabcolsep}{4pt}
\begin{tabular}{c|ccccc} 
\toprule
Type & Method &$\#$Params (M) &  Throughput & CIFAR10(\%)  & CIFAR100(\%) \\
\midrule
\emph{CNNs} & Pyramid-164-48  & 1.7 &  3715.9 & 95.97 & 80.70 \\
& WRN28-10    & 36.5 &   1510.8 & 95.83 & 80.75 \\
\midrule
& Deit-T  &5.3 & 1905.3 & 88.39 & 67.52 \\
& Deit-S &  21.3 &   734.7 & 92.44 & 69.78 \\
& Deit-B &  85.1 &  233.7 & 92.41 & 70.49 \\
\cmidrule{2-6}
& PVT-T & 12.8 &   1478.1 & 90.51 & 69.62 \\
& PVT-S  &24.1  &  707.2 & 92.34 & 69.79 \\
& PVT-B  &60.9  &   315.1 & 85.05 & 43.78 \\
\cmidrule{2-6}
\emph{Trans-} & Swin-T & 27.5 &  1372.5 & 94.46 & 78.07 \\
\emph{formers} & Swin-S  & 48.8 &    2399.2 & 94.17 & 77.01 \\
 & Swin-B  & 86.7 &  868.3 & 94.55 & 78.45 \\
 \cmidrule{2-6}
 & Nest-T & 6.2 &   627.9 & 96.04 & 78.69 \\
& Nest-S  &23.4 &   1616.9 & 96.97 & 81.70 \\
 & Nest-B  &90.1 &  189.8 & 97.20 & 82.56 \\
  \cmidrule{2-6}
& \textbf{Our-T} & 10.2 &  1700.2 & \textbf{97.00}  & \textbf{81.80} \\
& \textbf{Our-S} &   36.2 &  807.6 & \textbf{97.64}  & \textbf{83.66} \\
& \textbf{Our-B}&  115.8 &  276.8 &\textbf{97.75} & \textbf{84.70} \\
\bottomrule 
\end{tabular}
\caption{Classification accuracy on CIFAR10 and CIFAR100. 
 The number of parameters (in millions), and inference throughput (images per second) on a single GPU are shown. 
\label{tab:cifar}
}
\end{table*}

\vspace{0.2cm}
\noindent\textbf{ImageNet}\quad
In Tab.~\ref{tab:imagenet}, we consider a comparison of our method with baselines on the ImageNet dataset, which consists of a much larger number of higher-resolution ($224 \times 224$) images. Since the image resolution is $224 \times 224$, a fixed partition of the image into patches of size $S \times S$ results in $\frac{224}{S} \cdot \frac{224}{S}$ patches. 

As the attention mechanism is quadratic in the number of tokens $B$, $S$ is either chosen to be large (ViT, DeiT), or a subset of patches is used as tokens (Nest, Swin). 

{In contrast, our method considers, for each patch, $T=10$ possible local variants: (0,0), (1,0), (2,0), (3,0), (0,1), (0,2), (0,3), (1,1), (2,2), (3,3)}. As a result, it can correctly capture details from its local neighborhood. Our local attention layer results in quadratic computation only in the number of variants. That is, the cost is $O(B \cdot T^2)$. The cost of subsequent global attention layers is $O(B^2)$. Therefore, as long as $T \leq \sqrt{B}$, our method results in lower or equal computation cost, and it also considers local variants of each patch. As can be seen in Tab.~\ref{tab:imagenet}, this results in a superior performance of our method in comparison to baselines for two different model sizes. 

\begin{table}[t]
\centering
\renewcommand{\arraystretch}{0.95}
\setlength{\tabcolsep}{4pt}
\begin{tabular}{c|ccccc} 
\toprule
Method & $\#$Prms &  GFLOPs & Through & Acc(\%) \\
\midrule
ResNet-50 &  25 & 3.9 & 1226.1 & 76.2 \\
ResNet-101 &  45 & 7.9 & 753.6 & 77.4 \\
RegNetY-4GF  & 21 & 4.0 & 1156.7 & 80.0 \\
RegNetY-8GF  & 39 &  8.0 & 591.6 &81.7 \\
\midrule
Deit-T  & 5.7  & 1.3 & 2536.5  &72.2   \\
 PVT-T  & 13.2 & 1.9 & - &75.1   \\
\textbf{Our-T}  & \textbf{9.9} &  3.1  & 797.2 & \textbf{76.0}  \\
\cmidrule{1-5}
PVT-S &   
24.5 & 3.8 & - &79.8    \\
Deit-S &  22.0 & 4.6 & 940.4 & 79.8  \\
Swin-T  & 29.0 & 4.5 & 755.2 &81.3    \\
Nest-T  & 17.0  &5.8 & 633.9  & 81.5 \\
 CVT-T  & 20.0 & 4.5 & - & 81.6 \\
 CrossViT-S & 26.7 & 5.6 & - & 81.0 \\
 T2T-ViT-14 & 22.0 & 5.2 & - & 81.5 \\
 TNT-S & 23.8 & 5.2 &- & 81.3 \\
  \textbf{Our-S} & \textbf{22.0} & 7.6 &298.9 &  \textbf{82.2}     \\
\bottomrule
\end{tabular}

\caption{Classification accuracy on the ImageNet validation set. Number of parameters (in millions), GFLOPs, and inference throughput (Through) in images per second, on single GPU are shown.  Throughput is given for baselines for which this value has been reported. Top part - CNN based methods, middle (resp. bottom) - methods comparable in number of parameters to Our-T (resp. Our-S). 
\label{tab:imagenet} }
\end{table}

\subsection{Ablation study}
\label{sec:ablation}

\begin{table}
\centering
\begin{tabular}{ccccccc} \\ 
\toprule
\thead{Method} & \thead{Model \\ Variant} & \thead{$\#$Shif \\ -ting} & \thead{Conv.\\ Varia \\ -tions} &  \thead{Pos. \\ Emb \\ -ed.} & \thead{$\#$Par\\ -ams\\(M)}  & \thead{CIFAR10\\(\%)} \\
\toprule
\thead{Baseline} & \thead{nested \\ -T} & 0 & - & - & 6.2  & 96.04 \\
 \midrule
 \thead{Ours w/o \\ shifting}   & A & 0 & - & - & 6.8   & 96.50    \\
\midrule
& B & 9 & \checkmark & \checkmark & 8.4  & 96.70 \\
 & C & 18 & \texttimes & \texttimes & 6.8 & 96.69  \\
\thead{Ours} & D & 18 & \texttimes & \checkmark & 10.1  & 96.77 \\
 & E & 18 & \checkmark & \texttimes & 7.6 & 96.85 \\
 & Full & 18 & \checkmark & \checkmark & 10.2  & 97.00 \\
\bottomrule
\end{tabular}
\caption{Ablation analysis for examining the contribution of: (1) shifting variants, (2) number of shifting variants, (3)  shifting variants generated using learned convolutional filters, and, (4) different positional embeddings for each shifting variant.}
\label{tab:ablation}
\smallskip
\smallskip
\centering
\begin{tabular}{cccc} \\ 
\toprule
\thead{Variant} & \thead{Local \\ Attention} & \thead{Global \\ Attention}  & \thead{CIFAR10\\(\%)} \\
\toprule
I & Swin block~\cite{ze2021swin} & \checkmark & 96.3 \\
II & CvT~\cite{wu2021cvt} & \checkmark & 96.0 \\
Full &  \checkmark & \checkmark &  97.00 \\
\bottomrule
\end{tabular}
\caption{Analysing the contribution of each component.}
\label{tab:ablation_components}
\end{table}

An ablation study is performed for our method. The results are summarized in Tab.~\ref{tab:ablation}. To demonstrate the superiority of our method, we consider the current state of the art as baseline~\cite{zhang2021aggregating}. First, we examine the importance of shifting, i.e. introducing shifting variants as input to the model and attending them with a local attention layer. By removing the use of shifting variants and local attention layers, our method is simplified to only applying hierarchical global attention, as described in Sec.~\ref{sec:localglobal}. As can be seen (variant A), the tiny setting our method surpasses the baseline even without the use of shifting variants. Comparing the shiftless variant A to our full method (variant Full) demonstrates the performance gain achieved by adding shifting variants.

Next, we analyzed the effect of shifting the image with different number of variants (\#Shifting). As can be seen (variants A to B to F), increasing the number of shifting variants improves the model's performance. To reduce the number of shifts, we focus on shifts that are either on the horizontal or vertical lines, i.e, the 9 shifts: $(0,0),(0,1),(0,2),(1,0),(2,0),(-1,0),(-2,0),(0,-1), \allowbreak (0,-2)$, or a subset of the shifts in which we include diagonal lines as well, i.e., the 18 shifts out of the possible 25: $(0,0),(1,0),(0,1),(-1,0),(0,-1),(2,0),(0,2),(-2,0),\allowbreak (0,-2),  (1,1),(1,2),(1,-1),(-1,1),(-1,2),(-1,-1), \allowbreak (2,1),(2,2),(2,-1)$.

In Sec.~\ref{sec:shifting} we describe the initial pre-processing step of embedding the patches and shifting variants, which is applied before the local attention block. We now compare two approaches to shifting: (i) constructing $T$ shifting variants as a pre-processing step, using image translation with reflection padding and (ii) passing the original image as input to the network, and applying $T$ convolutional layers as in Sec.~\ref{sec:shifting} (as is done in our method). In Tab.~\ref{tab:ablation},  ``Conv. Variations'' indicates applying (ii) as opposed to (i). As can be seen, comparing variant C with variant E, and comparing variant D with our full method (variant Full), generating shifting variants using learned convolutional layers improves the model performance. 
Next, we checked whether it is necessary to add different positional embeddings to each shifting variant rather than simply learning one set of positional embeddings for all $T$ variants. As can be seen, comparing variant C with variant D and variant E with ur full method (variant Full), there is a trade-off between parameters and accuracy. Adding positional embeddings for each variant improves the performance, but the number of parameters increases. In our experiments we apply different positional embeddings to each shifting variant. 
As can be seen, our complete method, variant Full, outperforms the baseline. Furthermore, as can be seen in Tab.~\ref{tab:cifar}, our tiny model (Our-T) outperforms the small model of the baseline methods (Deit-S, PVT-S, Swin-S and Nest-S). 

In Tab.~\ref{tab:ablation_components}, we perform an additional ablation on the necessity of our local attention block. 
We consider variant I, in which we replace our local attention block with Swin's transformer block~\cite{ze2021swin}. While this variant achieves $96.3\%$ accuracy on CIFAR10, Swin-T achieves $94.5\%$  (see Tab.~\ref{tab:cifar}), demonstrating the importance of local attention for early global integration.  In variant II, we replace our local attention block with the CvT local attention procedure~\cite{wu2021cvt}, and make two modifications in our local attention block: (1) replace our patch embedding with Convolutional Token Embedding with overlapping patches (2) replace the linear projection of the keys, queries and values in the self attention blocks with Convolutional Projection. Our method outperforms both variants. 

\section{Limitations and Environmental Impact}

We note that in order to train our models, as for other vision transformers, many GPU hours are required. This may result in a high environmental impact, especially in terms of carbon footprint.

\section{Conclusions}

While convolutional layers typically employ small strides, resulting in heavily overlapping patches, the token-based approach of recent transformer-based techniques has led to a grid view of the input image. This leads to the loss of the translation invariance that played a major role in the development of neural networks in computer vision, as well as in the study of biological vision models~\cite{NIPS2009_2ab56412}.

In this work, we reintroduce the locality bias of CNNs into a transformer-based architecture, already at the very first layer.  This has the benefit of being able to model fine-detailed local correlations in addition to the coarse-detail global correlations for low-level features, which transformers model well. We employ two types of attention layers. The local attention layer models the correlation of a patch with its local shifting variants, thus modeling fine-grained correlations. The global attention layer, applied in a pyramidal manner, with decreasing input resolution, models long-range correlations. The use of local-global attention layers as opposed to a single attention layer is crucial for introducing the desired locality bias and capturing the correlation between neighbouring local shifts of each patch. This is especially useful for smaller datasets, with low-resolution images, such as CIFAR10/CIFAR100. Nevertheless, our method also scales well to large images, such as those of ImageNet. 

We demonstrate the superiority of our method on both small-resolution inputs of $32 \times 32$ (CIFAR10/CIFAR100) and larger-resolution inputs of $224 \times 224$ (ImageNet). Our method achieves superior accuracy to other convolutional and transformer-based state-of-the-art methods, with a comparable number of parameters.

\section*{Acknowledgments}
This project has received funding from the European Research Council (ERC) under the European Unions Horizon 2020 research and innovation programme (grant ERC CoG 725974).

{\small
\bibliographystyle{ieee_fullname}
\bibliography{egbib}
}

\end{document}